\documentclass[a4paper,twoside]{article}
\usepackage[authoryear,round,longnamesfirst]{natbib}
\usepackage{epsfig}
\usepackage{subcaption}
\usepackage{calc}
\usepackage{amssymb}
\usepackage{amstext}
\usepackage{amsmath}
\usepackage{amsthm}
\usepackage{multicol}
\usepackage{pslatex}
\usepackage{apalike}
\usepackage{booktabs}
\usepackage{comment}
\usepackage{SCITEPRESS}     

\begin{document}

\title{DLVGen: A Dual Latent Variable Approach to Personalized Dialogue Generation}

\author{\authorname{Jing Yang Lee\sup{1}\orcidAuthor{0000-0002-3611-3512}, Kong Aik Lee\sup{2}\orcidAuthor{0000-0001-9133-3000} and Woon Seng Gan\sup{1}\orcidAuthor{0000-0002-7143-1823}}
\affiliation{\sup{1}School of Electrical and Electronic Engineering, Nanyang Technological University, Singapore}
\affiliation{\sup{2}Institute for Infocomm Research, A*Star, Singapore}
\email{jingyang001@e.ntu.edu.sg, kong\_aik\_lee@i2r.a-star.edu.sg, ewsgan@ntu.edu.sg}
}


\abstract{The generation of personalized dialogue is vital to natural and human-like conversation. Typically, personalized dialogue generation models involve conditioning the generated response on the dialogue history and a representation of the persona/personality of the interlocutor. As it is impractical to obtain the persona/personality representations for every interlocutor, recent works have explored the possibility of generating personalized dialogue by finetuning the model with dialogue examples corresponding to a given persona instead. However, in real-world implementations, a sufficient number of corresponding dialogue examples are also rarely available. Hence, in this paper, we propose a Dual Latent Variable Generator (DLVGen) capable of generating personalized dialogue in the absence of any persona/personality information or any corresponding dialogue examples. Unlike prior work, DLVGen models the latent distribution over potential responses as well as the latent distribution over the agent's potential persona. During inference, latent variables are sampled from both distributions and fed into the decoder. Empirical results show that DLVGen is capable of generating diverse responses which accurately incorporate the agent's persona.}

\onecolumn \maketitle \normalsize \setcounter{footnote}{0} \vfill
\newcommand{\indep}{\perp \!\!\! \perp}
\section{\uppercase{Introduction}}
\label{sec:introduction}
Personalized dialogue generation refers to the task of generating coherent, fluent dialogue consistent with a specific persona or personality. The generation of personalized dialogue is key to achieving natural, human-like dialogue. Conventionally, this task requires a representation of the interlocutor's persona or personality. This representation can either be explicitly provided in the form of a textual persona description, which consists of multiple persona statements, or via personality related metadata such as gender, age etc. Recent work have also explored inferring the persona statement directly from the dialogue context. The persona/personality representation, along with the dialogue context, is then used to condition the decoder.



Due to the difficulty of crafting persona descriptions and personality representations in a practical setting,  the Persona Agnostic Meta-Learning (PAML) framework \citep{madotto-etal-2019-personalizing} and the Multi-Task Meta-Learning (MTML) framework \citep{lee-etal-2021-generating}  were proposed. Both PAML and MTML involved pretraining the model via meta-learning and finetuning with dialogue examples corresponding to the dialogue agent's persona. Similarly, \citep{DBLP:journals/corr/abs-1911-04700} also introduced a pretraining and finetuning approach for persona-sparse dialogue which features an attention routing structure and a learned persona attribute embedding. However, even though these methods do not require the explicit provision of any persona/personality information, multiple dialogue examples corresponding to the dialogue agent's persona is still needed to finetune the model. This is impractical as corresponding dialogue examples are rarely available in real-world implementations. Hence, in this paper, we propose a novel Dual Latent Variable Generator (DLVGen) for personalized dialogue generation, which does not rely on any persona/personality information or on any corresponding dialogue examples. Instead, DLVGen generates the response given only the dialogue context.

A common issue with regard to personalized dialogue agents is the low response diversity. Prior work address this issue by exploring the application of latent variable models, specifically the Conditional Variational Auto Encoder (CVAE) \citep{NIPS2015_8d55a249}. These approaches involved modelling the potential dialogue responses as a latent Gaussian distribution, which improves diversity due to the introduced stochasticity. Typically, the persona or personality information and dialogue history are used to generate the latent Gaussian distribution. During inference, the persona or personality vector, the dialogue context, along with a latent variable sampled from the latent distribution, are passed to the decoder for response generation. The Persona-CVAE framework \citep{Song2019ExploitingPI} aims to generate responses which incorporate information from the provided textual persona description, which is incorporated into the decoding process via a multi-hop attention mechanism. On the other hand, the Persona-Aware Variational Response Generator (PAGenerator) \citep{wu-etal-2020-guiding} relies on user embeddings trained concurrently with the CVAE. In addition to the CVAE, the basic Variational Auto Encoder (VAE) and the Wasserstein Auto Encoder (WAE) have also been applied to personalized dialogue generation in a similar manner \citep{chan-etal-2019-modeling}. The Common Sense and
Persona Aligned Chatbot (COMPAC) \citep{majumder-etal-2020-like}, on the other hand, relies on a latent variable to select the appropriate persona information from a set of expanded persona descriptions.

Unlike the aforementioned models, DLVGen involves generating two latent Gaussian distributions: the latent distribution over the agent's potential persona (defined by multiple persona statements in the persona description), and the latent distribution over potential dialogue responses. While it has been established that modelling the latent distribution over responses would increase response diversity, we hypothesize that modelling latent distribution over the agent's potential persona would result in responses which incorporate a range of potential persona information inferred from the dialogue context. We utilize the persona description only during training of a neural network that is tasked with generating the latent distribution over the agent's potential persona. To the best of our knowledge, this is the first attempt at modelling the latent distribution over the agent's \emph{potential} persona. For this paper, our contributions are three-fold:
\begin{enumerate}
    \item We propose the DLVGen framework which leverages persona descriptions only during training. Unlike prior frameworks, the proposed framework models both the latent distribution over potential dialogue responses as well as the latent distribution over the agent's potential persona.
    \item We introduce a variance regularization technique which involves maximizing or minimizing the variance of the distribution over the agent's potential persona and the distribution over potential responses respectively. In particular, evaluation results reveal that minimizing the variance of the distribution over potential responses improves the persona consistency of the generated responses.
    \item We present a selection framework based on lexical diversity to select the final response from a pool of generated responses. Generally, we find that the response selected via lexical diversity selection demonstrate greater persona consistency and diversity.
\end{enumerate}

\section{METHODOLOGY}
\subsection{Task Definition}
In this paper, we will tackle the task of generating personalized dialogue responses in the absence of any persona/personality information or corresponding dialogue examples. 
Hence, during inference, the model is expected to generate personalized dialogue given only the dialogue context, which consists of all previous utterances in the conversation. In other words, the model is expected to infer the agent's persona from only the dialogue context. For this task, the generated response is expected to be diverse as well as persona consistent i.e., accurately reflect the agent's persona. However, in cases where no persona information can be inferred from the dialogue context, the responses generated should aim to be persona neutral. This means that, as far as possible, responses should not contain any personal information which could potentially contradict the agent's persona.

For our experiments, we utilize the ConvAI2 dialogue corpus. We chose the ConvAI2 corpus as the textual persona descriptions, which consists of multiple persona statements, can be used during training. Also, the available persona descriptions allows us to evaluate the amount of persona information that is accurately reflected in the generated response relatively easily.

\subsection{Dual Latent Variable Generator}

\begin{figure*}[ht]
 \centering
 \includegraphics[height=2.2in,width=1.0\textwidth]{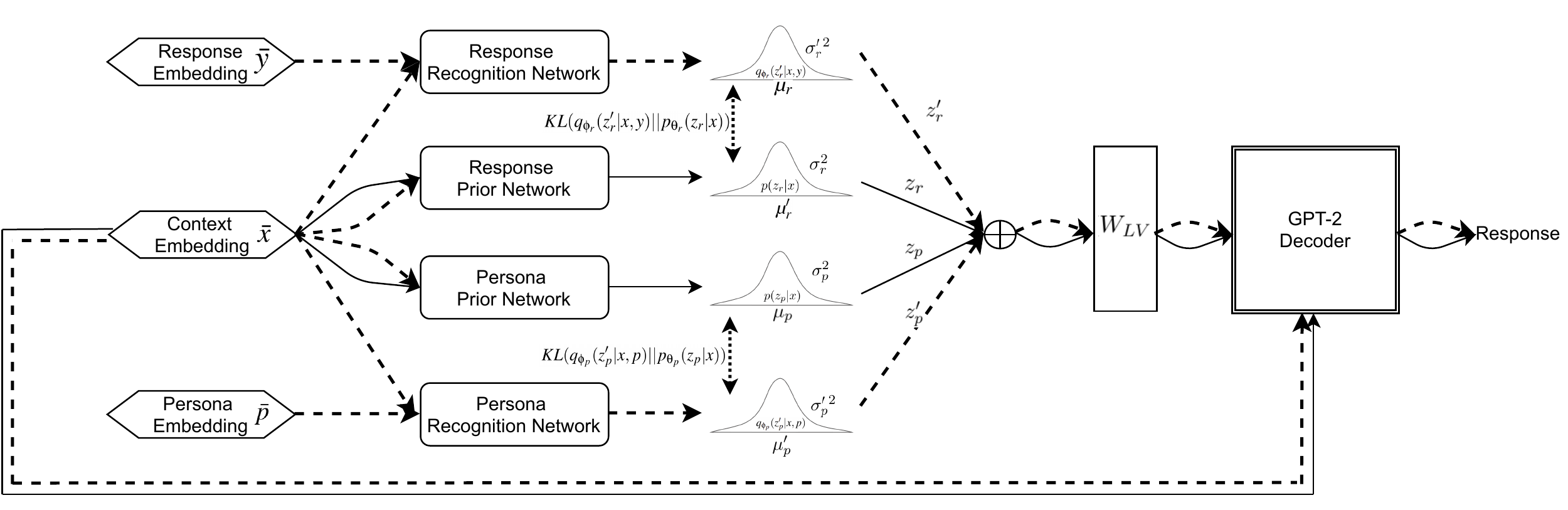}
 \caption{Flowchart depicting the architecture of the DLVGen model. The dashed lines and solid lines represents the connections that occur only during training and inference respectively. The dotted bidirectional connections indicate the KL divergence computation. $\oplus$ represents the concatenate operation.}
\end{figure*}

Essentially, the proposed Dual Latent Variable Generator (DLVGen) framework requires learning two distinct networks to model the latent distribution over potential dialogue responses as well as the latent distribution over the agent's potential persona. During inference, latent variables are sampled from both distributions and fed to the decoder, which consists of a GPT-2 pretrained language model \citep{radford2019rewon}.  Randomly sampling the latent distribution over responses would improve response diversity due to the stochasticity introduced by the latent variable. Similarly, randomly sampling the latent distribution over the agent's persona would incorporate a wide range of potential persona information (inferred from the dialogue context) in the generated response. The induced stochasticity would also reduce the likelihood of the model incorporating the same persona information in multiple responses, further contributing to the overall response diversity. In our approach, the persona statements describing each interlocutor is only utilized during training. An overview of the model is provided in Figure 1.

In our discussion, $x$ represents the dialogue context which consists of all the prior utterances in the dialogue history. $p$ refers to all statements in the provided textual persona description. $y$ denotes the dialogue response. During inference, the dialogue response is generated based on the dialogue context $x$, the latent variable sampled from the distribution over the agent's potential persona $z_{p}$, and the latent variable sampled from the distribution over potential dialogue responses $z_{r}$. Since $z_{p}$ and $z_{r}$ each represent different aspects of the generated response ($z_{p}$ encompasses the persona information and $z_{r}$ captures information regarding the flow of the dialogue), we will regard $z_{p}$ and $z_{r}$ as independent i.e., $z_{p} \indep z_{r}$. The persona information and dialogue flow information, which are both derived from the dialogue context, are disentangled via the prior and recognition networks introduced in the remainder of this section. Hence, by assuming $z_{p}$ and $z_{r}$ are independent random variables, the generation process can be expressed via the following conditional distribution $p(y,z_{p},z_{r}|x) = p(y|x,z_{p},z_{r})p(z_{p}|x)p(z_{r}|x)$.  We also assume that $z_{p}$ and $z_{r}$ can be modelled by multivariate isotropic Gaussian distributions:
\begin{equation}
p(z_{p}|x) = N(\mu_{p},\sigma_{p}^{2}\textbf{I})
\end{equation}
\begin{equation}
p(z_{r}|x) = N(\mu_{r},\sigma_{r}^{2}\textbf{I})
\end{equation}
where $\mu_{p}, \mu_{r}$ and $\sigma_{p}^{2}, \sigma_{r}^{2}$ refer to the mean and variance of the distribution over the potential persona and the distribution over potential responses respectively. Hence, to approximate $p(z_{p}|x)$ and $p(z_{r}|x)$, we define 2 prior networks $p_{\theta_{p}}(z_{p}|x)$ and $p_{\theta_{r}}(z_{r}|x)$, which are single-layer Multi-Layer Perceptrons (MLPs) parameterized by $\theta_{p}$ and $\theta_{r}$. $p_{\theta_{p}}(z_{p}|x)$ and $p_{\theta_{r}}(z_{r}|x)$ are termed the persona prior network and the response prior network respectively. Before being fed to the prior networks, a GPT-2 pretrained model \citep{radford2019rewon} is used to obtain $\bar{x}$, $\bar{y}$ and $\bar{p}$, the sequence embeddings of the dialogue context $x$, response label $y$ and persona descriptions $p$ respectively. Utilizing the persona and response prior networks, the following means $\mu_{p},\mu_{r}$ and variances $\sigma_{p}^{2},\sigma_{r}^{2}$ are derived as follows:
\begin{equation}
\begin{bmatrix} \mu_{p} \\log(\sigma_{p}^{2}) \end{bmatrix} = W_{p} \begin{bmatrix} \bar{x} \end{bmatrix} + b_{p}
\end{equation}
\begin{equation}
\begin{bmatrix} \mu_{r} \\log(\sigma_{r}^{2}) \end{bmatrix} = W_{r} \begin{bmatrix} \bar{x} \end{bmatrix} + b_{r}
\end{equation}
where $W_{p}, W_{r}$ refer to the weight vectors and $b_{p},b_{r}$ refer to the bias vectors corresponding to the persona and response prior networks.

Additionally, to approximate the posterior of $p(z_{p}|x)$ and $p(z_{r}|x)$,  we introduce latent variables $z'_{p}$ and $z'_{r}$ as well as latent distributions $q(z'_{p}|x,p)$ and $q(z'_{r}|x,y)$, which are generated by 2 recognition networks $q_{\phi_{p}}(z_{p}'|x,p)$ and $q_{\phi_{r}}(z_{r}'|x,y)$ (defined as single-layer MLPs parametrized by $\phi_{p}$ and $\phi_{r}$) respectively. $q_{\phi_{p}}(z'_{p}|x,p)$ is termed the persona recognition network and $q_{\phi_{r}}(z'_{r}|x,y)$ is termed the response recognition network. Similarly, the sequence embeddings $\bar{x}$, $\bar{y}$ and $\bar{p}$ are fed to the recognition networks. The persona and response recognition networks define the approximate posterior $q(z'_{p}|x,p)$ and $q(z'_{r}|x,y)$ via generating the respective means $\mu'_{p}, \mu'_{r}$ and variances $\sigma'^{2}_{p}, \sigma'^{2}_{r}$:
\begin{equation}
\begin{bmatrix} \mu'_{p} \\log(\sigma'^{2}_{p}) \end{bmatrix} = W'_{p} \begin{bmatrix} \bar{x}\\\bar{p} \end{bmatrix} + b'_{p} 
\end{equation}
\begin{equation}
\begin{bmatrix} \mu'_{r} \\log(\sigma'^{2}_{r}) \end{bmatrix} = W'_{r} \begin{bmatrix} \bar{x}\\\bar{y} \end{bmatrix} + b'_{r}
\end{equation}
where $W_{p}', W_{r}'$ refer to the weight vectors and $b_{p}',b_{r}'$ refer to the bias vectors corresponding to the persona and response recognition networks. The KL divergence between the prior and recognition networks will be included in the final loss function. It should be noted that for the persona prior network and the persona recognition network, before being fed into the GPT-2 pretrained model to obtain the embedding, the utterances corresponding to the opposing interlocutor (the user) are masked. This prevents the generated Gaussian distribution from modelling the persona of the opposing interlocutor. 

To ensure end-to-end differentiability, latent variables are sampled using the reparameterization trick \citep{kingma2014autoencoding}. During training, the latent variables $z'_{p}$ and $z'_{r}$ are sampled from the persona and response recognition networks. However, during inference, the latent variables $z_{p}$ and $z_{r}$ are sampled from the persona and response prior networks. After sampling, the latent variables $z'_{p}$ and $z'_{r}$ (training) or $z_{p}$ and $z_{r}$ (inference) are concatenated and fed to a linear layer ($W_{LV}$) before being passed to the decoder, which constitutes a GPT-2 pretrained language model. The resultant representation is added to the GPT-2 input, which consists of the token embedding and the positional encoding, at every decoding step. The persona description $p$ is only used during training to learn the latent distribution over the the agent's potential persona. A simple figure depicting this process in provided in Figure 2.

Since maximizing the conditional log likelihood during training requires an intractable marginalization over $z_{p}$ and $z_{r}$, DLVGen is trained via the Stochastic Gradient Variational Bayes (SGVB) framework \citep{kingma2014autoencoding} by maximizing the variational lower bound. Also, the Bag-of-Words (BoW) loss \citep{zhao-etal-2017-learning} is incorporated into the loss function to address the vanishing latent variable problem. Essentially, SGVB involves maximizing the variational lower bound of $p(y|x)$: 
\begin{equation}
\setlength{\jot}{1pt}
\begin{split}
&L(\theta_{p},\theta_{r},\phi_{p},\phi_{r},x,y,p,z_{p},z_{r}) = \\ &E_{q_{\phi_{p}}(z_{p}'|x,p);q_{\phi_{r}}(z_{r}'|x,y)}[log\:p(y|x,z_{p},z_{r})] \\  
&- KL(q_{\phi_{p}}(z'_{p}|x,p)||p_{\theta_{p}}(z_{p}|x))\\ 
&- KL(q_{\phi_{r}}(z'_{r}|x,y)||p_{\theta_{r}}(z_{r}|x))\\
&+E_{q_{\phi_{p}}(z_{p}'|x,p);q_{\phi_{r}}(z_{r}'|x,y)}[log\:p(y_{bow}|x,z_{p},z_{r})] 
\end{split}
\end{equation}
where $KL(\cdot)$ refers to the KL divergence and $y_{bow}$ represents the response bag-of-words.

\begin{figure}
 \noindent\makebox[\linewidth]{\includegraphics[width=0.7\linewidth]{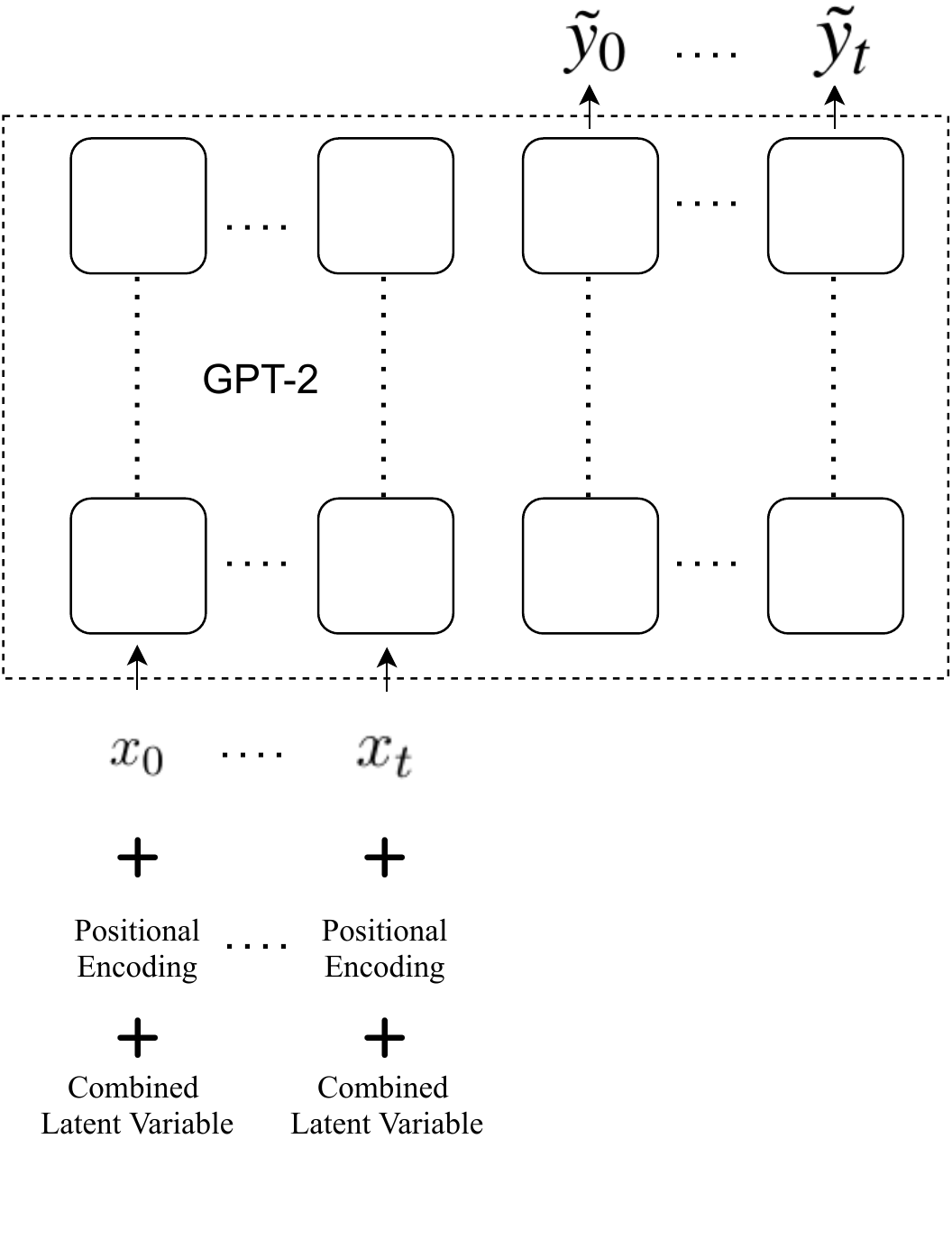}}
 \caption{Diagram depicting the GPT-2 decoder during inference. After the latent variables are concatenated and fed into a linear layer $W_{LV}$, the resultant combined latent variable is added to the positional encoding and token embedding at every decoding step. $x_{0}$ and $x_{t}$ represent the first and last token embedding of the dialogue context. $y_{0}$ and $y_{t}$ represent the first and last token of the generated response.}
\end{figure}

\subsection{Variance Regularization}
Additionally, we introduce a novel regularization technique based on the variances of distributions generated by the prior networks i.e., $p(z_{p}|x)$ and $p(z_{r}|x)$. Essentially, we either maximize the variance of the latent distribution over the agent's potential persona $\sigma_{p}^{2}$, or minimize the variance of the latent distribution over potential responses $\sigma_{r}^{2}$. Specifically, this is realized by either adding the euclidean norm of the log-variance of $p(z_{r}|x)$, $log(\sigma_{r}^{2})$, or the inverse of the euclidean norm of the log-variance of $p(z_{p}|x)$, $log(\sigma_{p}^{2})$. The inverse of the euclidean norm of the log-variance can also be interpreted as the precision of the distribution. The variance regularization terms can be expressed as:
\begin{equation}
R_{r} = \lambda_{r}\:\lVert log(\sigma_{r}^{2}) \lVert
\end{equation}
\begin{equation}
R_{p} = \lambda_{p}\lVert \frac{1}{log(\sigma_{p}^{2})} \lVert
\end{equation}
where $R_{r}$ refers to the loss term which minimizes the variance of the distribution over potential responses, and $R_{p}$ refers to the loss term which maximizes the variance of the distribution over the agent's potential persona. $\lambda_{r}$ and $\lambda_{p}$ are penalty terms to be tuned. Then, the 2 variance regularization terms can be individually added to the loss function. Alternatively, both terms can be added to the loss function to form the final loss function $\bar{L}$:
\begin{equation}
\setlength{\jot}{1pt}
\begin{split}
\bar{L}&(\theta_{p},\theta_{r},\phi_{p},\phi_{r},x,y,p,z_{p},z_{r}) =\\  &L(\theta_{p},\theta_{r},\phi_{p},\phi_{r},x,y,p,z_{p},z_{r})\\
&+R_{r}(\sigma_{r}^{2}) +R_{p}(\sigma_{p}^{2})
\end{split}
\end{equation}
By maximizing the variance of the latent distribution over the agent's potential persona, we hope to encourage the persona prior network to model a wider range of potential persona information, which would further improve response diversity. On the other hand, by minimizing the variance of the latent distribution over potential responses, we aim to constrain the variability of the sampled random variables, improving the contextual coherence of the generated responses.

\subsection{Lexical Diversity Selection}
Lexical diversity is a measure of how many different lexical words appear in a given text. Unlike filler words such as `and' or `the', lexical words are words that convey information or meaning (`cat',`computer',`sky' etc.) .  By sampling the Gaussian distributions $N$ times, $N$ different responses can be generated. Hence, we introduce a method of selecting the best response from the pool of $N$ generated responses. Intuitively, we assume that lexically diverse sentences make for more informative responses. Hence, to quantify the lexical diversity of the generated responses, we utilize the Measure of Textual Lexical Diversity (MTLD) and the Moving Average Type Token Ratio (MATTR) scores \citep{doi:10.1044/2015-JSLHR-L-14-0280}. Obtaining the MTLD involves computing the Token-Type Ratio (TTR) for sequentially larger segments of the sentence. For each segment, when the TTR drops below a predefined threshold $h$, a count is incremented by 1. Then, the total number of tokens in the generated response is divided by the final count. The final MTLD is obtained by averaging the forward and backward MTLD scores. The MATTR, on the other hand is based on averaging the TTR of successive segments of the generated response with a fixed window size $w$. After the computing the MATTR and MTLD, we select the response with the highest combined score. This is summarized in the following expression:
\begin{equation}
\underset{r\in R}{\operatorname{argmax}} (0.1M_{1}^{h}(r) + M_{2}^{w}(r))
\end{equation}
where $M_{1}^{h}$ and $M_{2}^{w}$ refer to the functions for calculating the MTLD score (with threshold $h$) and MATTR (with window size $w$) score respectively. $R$ refers to the pool of generated responses. The MTLD score was penalized by a factor of 0.1 to prevent it from overwhelming the MATTR score. The proposed lexical diversity selection method is relatively straightforward to implement and can be applied on top of any latent variable dialogue model.
\section{EXPERIMENT}
\subsection{Corpus}
We evaluate our approach on the ConvAI2 personalized dialogue corpus \citep{dinan2019second}. Since the ConvAI2 corpus is based on the PersonaChat corpus \citep{zhang-etal-2018-personalizing}, both corpora are structurally similar. Both corpora provide personalized dialogues as well as persona descriptions in the form of several sentences. However, the ConvAI2 corpus features numerous crowd-sourced rewrites and rephrases, which increases the overall task difficulty. In ConvAI2, the training set $\textbf{D}_{train}$ contains 17,878 dialogues from 1155 unique personas, while the test set (validation set is used as the test set in our experiments as actual test set is hidden) $\textbf{D}_{test}$ contains 1000 dialogues from 100 unique personas.
\subsection{Implementation}
For our experiments, we utilize the PyTorch, ParlAI, and HuggingFace libraries. During training, the Adam optimizer (learning rate = 0.0001) is used with a batch size of 16. The size of latent variables $z_{p}$, $z_{r}$, $z'_{p}$, $z'_{r}$ are fixed at 256. We utilize the GPT-2 pretrained language model \citep{radford2019rewon} from HuggingFace to obtain the context, persona and response embeddings. The decoder also consists of a GPT-2 model (12 layers, 768 dimensional hidden state, 12 heads, 117 million parameters). 

For variance regularization, the values of $\lambda_{r}$ and $\lambda_{p}$ was set to 0.5 and 1.0 respectively. For the MTLD and MATTR computation during lexical diversity selection, $h$ and $w$ were fixed at 0.72 and 4 respectively. Responses are generated via beam search with a beam size of 3. $N$ is also set to 3.
\subsection{Baselines}

To evaluate our proposed model on the task described in section 2.1, we implement the following baselines:

\begin{enumerate}
    \item \textbf{PAML} Following \citet{madotto-etal-2019-personalizing}, we pretrain a standard transformer with PAML. However, using the ConvAI2 corpus, the model was applied to the task defined in section 2.1 instead (PAML was originally evaluated on the PersonaChat corpus), which involved directly evaluating the pretrained transformer without further finetuning.
    \item \textbf{MTML} Following \citet{lee-etal-2021-generating}, we pretrain a standard transformer with MTML ($\alpha=0.8$). Similarly, using the ConvAI2 corpus, the model was applied to the task defined in section 2.1 (MTML was also originally evaluated on the PersonaChat corpus), which involved directly evaluating the pretrained transformer without further finetuning.
    \item \textbf{GPT-2} Similar to TransferTransfo \citep{DBLP:journals/corr/abs-1901-08149}. However, instead of GPT, GPT-2 was finetuned for dialogue generation. Since the model was applied to the task defined in section 2.1, the decoder input consists of only the dialogue context. The persona description was not utilized.
    \item \textbf{CVAE} Similar to the CVAE-based dialogue model proposed by \citet{zhao-etal-2017-learning}, which involves only generating and sampling from the latent distribution over responses.  However, our implementation utilizes the GPT-2 pretrained model instead of the bidirectional GRU during decoding as well as to generate the sequence embeddings. Additionally, since the model was applied to the task defined in section 2.1, the persona description was not used. Instead, the prior network generates the latent Gaussian distribution based on only the dialogue context, and the decoder input consists of only the sampled latent variable and the dialogue context.
    \item \textbf{DLVGen} A model which consists of our proposed DLVGen architecture with a GPT-2 decoder. We also implement additional variations of this model which include each of the variance regularization terms $R_{r}$ and $R_{p}$, as well as the combination of both terms. Additionally, we implement the lexical diversity selection (LS) for all variants.
\end{enumerate}

\subsection{Evaluation}
In our experiments, $N$ was set to 3 i.e., the Gaussian distributions were sampled 3 times to generate 3 responses. For the latent variable models (CVAE and DLVGen), the final automatic and human evaluation scores are obtained by averaging the individual scores from the $N$ responses generated from each dialogue example in $\textbf{D}_{test}$. 

\subsubsection{Automatic Metrics}

\begin{table*}[h]
\caption{\label{tab:table-name} Results on the ConvAI2 dialogue corpus. $R_{r}$, $R_{p}$ and $LS$ refer to the variance regularization terms presented in Equation 5 and 6, and the lexical diversity selection respectively.}
\centering
\begin{tabular}{@{}lcccccccl@{}}
\toprule
                  & C-score & Distinct-1 & Distinct-2     &\begin{tabular}[c]{@{}c@{}}Persona \\ Consistency\end{tabular} & Naturalness & Engagingness \\ \midrule
PAML             &-0.029     &0.089           &0.182                 &0.041                                                              &0.141          &0.064           \\
MTML             &-0.055      &0.118           &0.268                 &0.057                                                              &0.116             &0.083          \\
GPT-2             & -0.011   & 0.097           & 0.177                 & -0.079                                                               &\textbf{0.378}             & 0.114           \\
CVAE              & -0.046   & 0.152       & 0.422       & 0.012                                                           &0.227             &0.091            \\ \midrule
DLVGen              & 0.048    & 0.151       & 0.426       &0.105                                                                & 0.276            &0.128            \\
+LS             & 0.049    & 0.367       & 0.757       & 0.123                                                               &0.356             &0.117            \\
+$R_{r}$             & 0.075    & 0.171       & 0.474       & 0.226                                                                &0.325            & 0.104           \\
+$R_{r}$+LS          & \textbf{0.081}    & \textbf{0.390}       &\textbf{0.809}                                             &\textbf{0.269}     &0.367             &0.151            \\
+$R_{p}$             &0.024     &0.161        &0.448        & 0.093                                                               & 0.181            & 0.134           \\
+$R_{p}$+LS          &0.017     &0.370        &0.775        & 0.142                                                                & 0.262            & 0.148           \\
+$R_{r}$+$R_{p}$           &0.054     &0.171        &0.474        & 0.173                                                           & 0.279            & \textbf{0.174}           \\
+$R_{r}$+$R_{p}$+LS & 0.062    & 0.386       & 0.801        & 0.214                                                               & 0.369            & 0.145           \\ \bottomrule
\end{tabular}
\end{table*}
We evaluate the generated responses with the following automatic metrics:


\begin{enumerate}
    \item \textbf{Distinct 1 \& 2} Distinct-1 and 2 scores quantify the diversity of a generated response. The Distinct-1 and Distinct-2 scores accounts for the number of distinct 1-grams and 2-grams in the generated response respectively. A higher Distinct-1 or 2 score would indicate greater response diversity.
    \item \textbf{C-score}  The C-score \citep{madotto-etal-2019-personalizing} reveals the extent to which the persona is accurately reflected in the generated response. Essentially, it is generated via a BERT model, which is finetuned to indicate if the generated response entails, contradicts or is independent to each statement in the persona description. A higher C-score would indicate a larger amount of accurately incorporated persona information.
\end{enumerate}

\subsubsection{Human Evaluation}
We engaged 5 graduates to evaluate the various generated responses based on 3 criteria which are similar to the criteria used during human evaluation in prior work\citep{lee-etal-2021-generating, madotto-etal-2019-personalizing}:

\begin{enumerate}
    \item \textbf{Persona Consistency} The individuals were told to rate the responses in terms of the amount of persona information present in the generated response on a scale from -1 to 1. A score of -1 would indicate a contradiction with the corresponding persona description and a score of 1 would indicate an accurate incorporation of persona information.
    \item \textbf{Naturalness} The individuals were told to rate the generated responses according to human-likeness on a scale from -1 to 1. This criteria accounts for both the diversity and general fluency of the generated response. A score of 1 would be assigned to a perfectly fluent response that is indiscernible from a human response. A score of -1 would indicate an awkwardly phrased response with multiple grammatical and lexical errors.
    \item \textbf{Engagingness} Individuals were told to rate the generated responses in terms of how engaging the generated response was, or how well the response attempts to continue the conversation, on a scale from -1 to 1. A score of 1 would indicate a response which follows the logical flow of the dialogue and aims to continue the conversation. A score of -1 would be assigned to a laconic response that is contextually incoherent.
\end{enumerate}

\section{RESULTS \& DISCUSSION}

\begin{table*}[]
\caption{\label{tab:table-name} Dialogue responses generated by DLVGen+$R_{r}$+LS. $Ref$ indicates the response label. Since $N = 3$, $r_{1}, r_{2}, r_{3}$ refer to the 3 responses generated after sampling the latent distributions 3 times. The response selected by the lexical diversity selector is in bold.}
\centering
\begin{tabular}{@{}ll@{}}
\toprule
\multicolumn{2}{c}{\textbf{Persona}}                \\ \midrule
\multicolumn{2}{l}{i have been trying all types of food everywhere i go.}             \\
\multicolumn{2}{l}{hey there i'm 23 and i love food.}           \\
\multicolumn{2}{l}{i've been traveling the world for a years.} \\
\multicolumn{2}{l}{i also like to cook but i'm not very good at it.}             \\
\multicolumn{2}{l}{i own a yacht and i rent it out when i'm not using it.}                 \\ \midrule
\multicolumn{2}{c}{\textbf{Dialogue Context}}       \\ \midrule
User:                       & hello. how are you today ?               \\
Agent:                       & good today . just cooking some mexican food. i cooking but am not very good.                     \\
User:                       & oh i like mexican food , but my favorite food are cheeseburgers.                      \\
Agent:                       & i inherited some money and bought a yacht to travel, i try different foods traveling                     \\
User:                       & i help out at a soup kitchen since i grew up poor                    \\
\midrule
\multicolumn{2}{c}{\textbf{Responses}}              \\ \midrule
$Ref:$                          &  cheeseburgers are great , i try all kinds of foods everywhere i go , gotta love food.           \\
$r_{1}:$                          & i love to cook , i am a chef.                      \\
$r_{2}:$                         & i love to cook.\\
$r_{3}:$                         &\textbf{i love to travel , i have been to many countries.}                  \\\bottomrule
\end{tabular}
\end{table*}

\subsection{Results}
The automatic and human evaluation results attained by the models described in section 3.4 are displayed in Table 1. An example of dialogue responses generated by the best performing DLVGen variant (DLVGen+$R_{r}$+LS) is provided in Table 2.

\subsection{Discussion}
Based on the C-scores and Persona Consistency scores attained, it can be easily observed that our proposed model incorporates the corresponding persona information to a larger extent compared to PAML, MTML, GPT-2 and CVAE. All DLVGen variants achieved higher C-scores and Persona Consistency scores. However, while the addition of the variance regularization term $R_{r}$ generally improves the C-score/Persona Consistency score, the introduction of $R_{p}$ leads to a slight drop in the same metrics. We suspect that this is because a large variance would increase the chances of modelling incorrect or contradictory persona traits, which would negatively impact the persona consistency of the responses. DLVGen variant that achieved the highest C-score and Persona Consistency score was DLVGen+$R_{r}$+LS. Also, it can be observed that the DLVGen model variants with lexical diversity selection achieved better C-scores/ Persona Consistency scores. This indicates that lexically diverse responses tend to contain more persona information. 

It should be noted that the C-scores and Persona Consistency scores obtained are dependant on the length of the dialogue context. A longer dialogue context (greater number of utterances) would usually imply greater persona consistency and larger C-score. This is expected as a short dialogue context would not have sufficient persona information to be inferred by the persona prior network.  For these cases, we observe that there is a tendency for the model to incorporate a randomly generated persona, which could be due to the failure of the persona prior network to model any persona information from the context. As a result, the generated response might contradict the agent's persona and negatively impact the C-score as well as Persona Consistency score. Ideally, in such cases, the model should be able to recognize that there is insufficient persona information in the dialogue context, and the generated response should be largely persona neutral (section 2.1) while remaining contextual and fluent. Addressing this issue could be a direction for future research. A potential approach could involve estimating the uncertainty, which could be represented by the variance, of the persona prior network, and incorporating the uncertainty estimate in the decoding process. 

Upon closer inspection of the responses generated by each of model, we notice that out of the responses generated by GPT-2 and CVAE, few attempt to incorporate the persona information. Instead, a significant fraction of responses were relatively short and generic. On the other hand, responses generated by DLVGen are more likely to a attempt some form of persona incorporation. It should be highlighted that even though it is less likely for generic responses to achieve high C-scores, it is also less likely for them to be assigned negative C-scores. A potential avenue for future work could involve designing a metric which penalizes generic responses in addition to inaccurate responses.

From our experimental results, the GPT-2 base model attained noticeably lower Distinct 1 and 2 scores compared to the latent variable models. This follows previous work which reported an increase in response diversity due to the stochasticity introduced by the latent variable \citep{zhao-etal-2017-learning, Song2019ExploitingPI}. Furthermore, the DLVGen model variants with lexical diversity selection achieved better Distinct-1 and 2 scores compared to all other models. This is not surprising since lexical diversity would generally imply general diversity. The improvements on both diversity and persona consistency demonstrate the efficacy of our lexical diversity selection approach.

When it comes to the Naturalness, we can observe that the usage of the GPT-2 pretrained language model resulted in higher scores. Both PAML and MTML, which utilize standard Transformers, attained noticeably lower Naturalness scores compared to all other models. This is to be expected as pretrained language models such as GPT-2 have demonstrated high language understanding and generation capability even without any finetuning.  We can also observe that most of the latent variable models attained comparable Naturalness scores. The only exception is the inclusion of the variance regularization term $R_{p}$, which results in a lower Naturalness score. The GPT-2 base model, however, maintains an edge over the latent variable models in terms on Naturalness score. This could be attributed to the short, generic responses generated by the base model, which are relatively more fluent and human-like. On the other hand, when it comes to the Engagingness score, all implemented models achieved relatively poor performance. Relatively few responses attempt to continue the conversation. Instead, most responses are largely informative.

\section{RELATED WORK}
\subsection{Latent Variable Models}
Latent variable models are a category of models that involve inferring latent random variables from a group of observable variables \citep{doi:10.1146/annurev-statistics-060116-054017}. Latent variable models such as the Variational Auto Encoder (VAE) \citep{kingma2014autoencoding} and the Conditional Variational Auto Encoder (CVAE) \citep{NIPS2015_8d55a249} have been applied to the task of open-domain dialogue generation, where the potential dialogue responses are modelled as a latent Gaussian distribution \citep{li-etal-2020-optimus, Shen2018ImprovingVE, zhao-etal-2017-learning, 10.5555/3298023.3298047}. In addition to personalized dialogue generation (examples provided in the introduction), CVAEs have been applied to conditional dialogue generation tasks such as emotional dialogue generation  \citep{LIU2021106,asghar2020generating,Zhou2018MojiTalkGE} as well as topical dialogue generation  \citep{article}.

\subsection{\textbf{Personalized Dialogue}}
Over the past few years, there has been numerous publications exploring various approaches to the task of personalized dialogue generation. As mentioned in the introduction, there are numerous approaches to the task of personalized dialogue generation. A popular approach typically involves conditioning dialogue responses on the dialogue context in addition to textual persona descriptions \citep{lee-etal-2021-generating,icaart21,liu-etal-2020-impress, majumder-etal-2020-like, madotto-etal-2019-personalizing,DBLP:journals/corr/abs-1901-08149,Zheng_Zhang_Huang_Mao_2020, chan-etal-2019-modeling, Song2019ExploitingPI}. This approach focuses on generating responses which incorporate the provided persona information. Another approach involves incorporating personality or persona related metadata into the decoding process \citep{ijcai2018-595}. Some approaches also involve implicitly learning personality user embeddings \citep{DBLP:journals/corr/abs-1901-09672, wu-etal-2020-guiding, DBLP:journals/corr/Al-RfouPSSSK16,li-etal-2016-persona}. Another approach entails inferring the dialogue agent's personality directly from the dialogue history \citep{DBLP:journals/corr/abs-1911-04700, Su2019}. Typically, for this approach, the primary aim is to train the agent to mimic the dialogue style of the interlocutor.

\section{CONCLUSION}
In this paper, we introduced DLVGen, a dual latent variable model which models the potential responses and the agent's potential persona as latent Gaussian distributions. Through our experiments, we find that responses generated by DLVGen effectively incorporates persona information inferred from the dialogue context. We also introduced a variance regularization technique and lexical diversity selection method which improves the quality of the generated responses in terms of both persona consistency and human-likeness. However, an area for improvement is the relatively poor engagingness of the dialogue. Encouraging the generation of persona consistent, diverse yet engaging open-domain dialogue is a potential avenue for future research. A possible approach involves designing an objetcive function which explicitly accounts for the engagingness of the generated response. The dialogue model could then be trained on both objective functions via a multi-task learning framework.

\bibliographystyle{apalike.bst}
{\small
\bibliography{example.bib}}



\end{document}